\documentclass[10pt,twocolumn,letterpaper]{article}

\usepackage[pagenumbers]{preprint}

\usepackage{placeins}
\usepackage{fancyhdr}

\fancypagestyle{preprint}{
    \fancyhf{}
    \fancyhead[R]{\small\textit{Preprint}}
    \fancyfoot[C]{\thepage}
    
}

\definecolor{linkblue}{rgb}{0.21,0.49,0.74}
\usepackage[pagebackref,breaklinks,colorlinks,allcolors=linkblue]{hyperref}

\def\paperID{}
\def\confName{}
\def\confYear{}

\title{U-JEPA: A Predictive Framework for Multimodal Alignment and Pretraining}

\author{
Atul Belekar \quad
Remish Leonard Minz \quad
Prateek Mital\\
Hitachi India Research and Development Center\\
Bangalore, Karnataka, India\\
{\tt\small \{atul.belekar,remish.minz,prateek.mital\}@hitachi.co.in}
}

\begin{document}

\maketitle
\thispagestyle{preprint}
\begin{abstract}
In this work, we investigate whether multimodal alignment can emerge from predictive learning without explicit contrastive optimization. We introduce \textbf{U-JEPA}, a joint-embedding predictive architecture for end-to-end multimodal pretraining that learns aligned representations without contrastive objectives or negative samples. U-JEPA employs an exponential-moving-average (EMA) teacher–student framework with pair-specific prediction targets, encouraging alignment to emerge directly within modality-specific encoders. Crucially, the learned representations remain aligned and directly comparable after the training-specific components are discarded. We evaluate U-JEPA on image–text pretraining using CC3M and CC12M datasets. Compared with CLIP baselines, U-JEPA improves zero-shot image–text retrieval by up to 16\% relative, substantially reduces the modality gap, and achieves strong performance across zero-shot classification, vision benchmarks, and compositional understanding, while also supporting token-level late interaction for additional gains. These findings demonstrate that strong cross-modal alignment can emerge from predictive learning without explicit contrastive optimization, suggesting JEPA as a strong alternative to contrastive objectives for multimodal alignment and transferable representations.
\end{abstract}    
\begin{figure}[t]
    \centering
    \includegraphics[width=\linewidth]{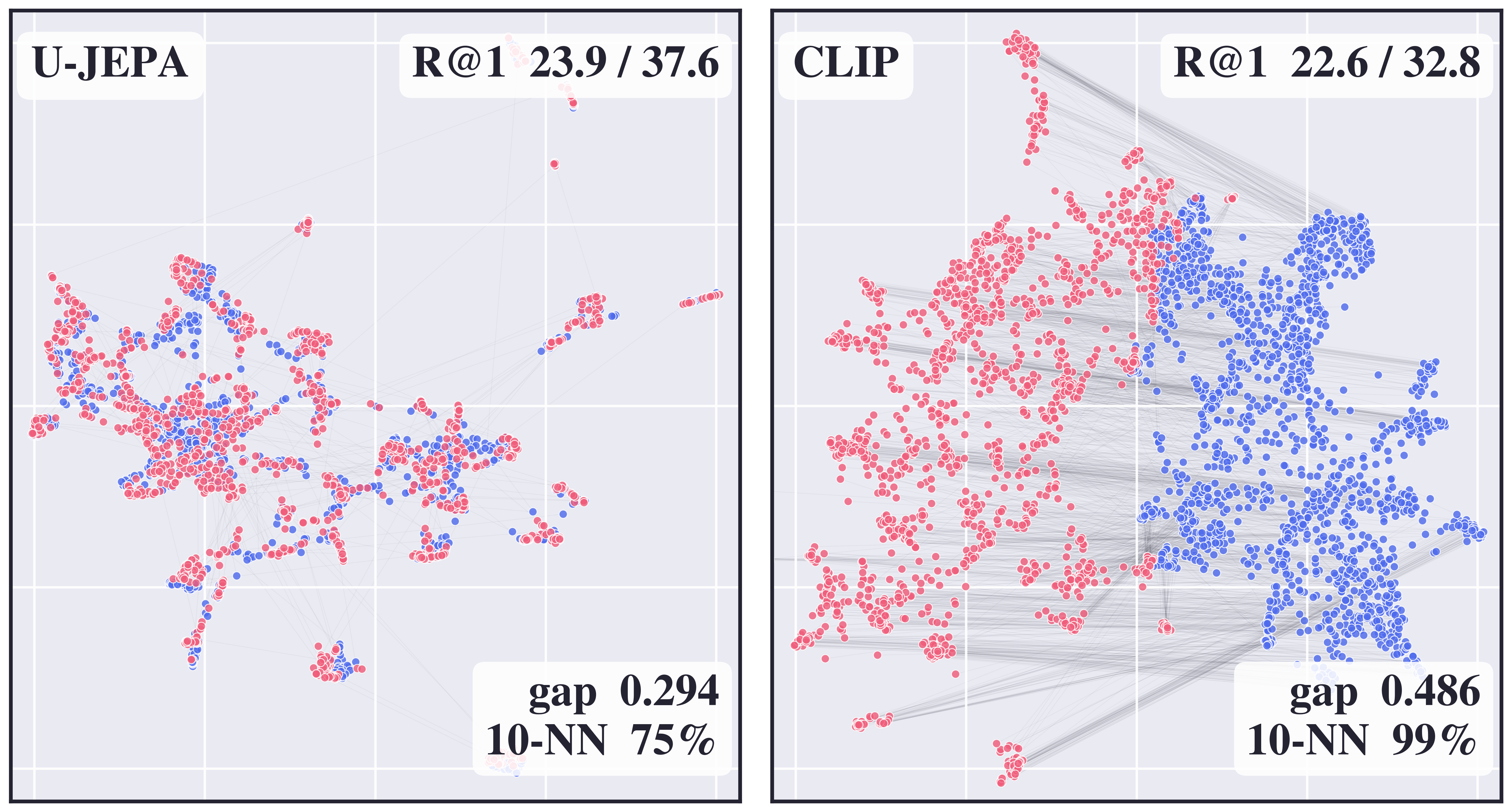}
    \caption{\textbf{Cross-modal embedding geometry of U-JEPA and CLIP.}
    UMAP visualization of image (blue) and text (red) embeddings on the COCO validation set; lines connect matched pairs. U-JEPA produces greater cross-modal overlap than CLIP, reducing the centroid-based modality gap measured using \citet{liang2022mindthegap} from 0.48 to 0.29 and 10-nearest-neighbor modality purity from 99.0\% to 75\% with improved retrieval. Both models are trained on CC12M dataset from scratch with matched training setup. (R@1: Text-to-Image/Image-to-Text)}. 
    \label{fig:embedding_geometry}
\end{figure}

\section{Introduction}

Learning a shared representation across modalities is a central challenge in multimodal learning. Contrastive methods such as CLIP, ALIGN, and SigLIP address it by distinguishing matched from mismatched image--text pairs \cite{radford2021clip,jia2021align,zhai2023siglip}. Yet even highly effective contrastive models can place images and text in systematically separated regions of their nominally shared space resulting in a persistent \emph{modality gap} \cite{liang2022mindthegap}. This separation raises a basic question about how multimodal representations should be learned: \emph{Does strong cross-modal alignment always need explicit comparison of matched vs unmatched pairs through contrastive discrimination during training?}

Self-distillation and joint-embedding predictive architectures (JEPAs) show that useful representations can be learned from latent prediction alone, without negatives or reconstruction objectives\cite{grill2020bootstrap,chen2021exploring,assran2023ijepa,bardes2024vjepa}. Across modalities, however, prediction does not imply alignment. A sufficiently expressive predictor can translate between incompatible modality specific spaces, minimize the objective, and leave the encoders no more aligned than before. This is the central obstacle: for efficient dual-encoder inference, cross-modal compatibility must reside in the encoders themselves, not in a predictor discarded after training.

We introduce \textbf{U-JEPA}, a joint-embedding predictive architecture that induces cross-modal alignment directly within modality-specific encoders, without contrastive discrimination or negative samples. Given a paired observation $(X_A, X_B)$, separate student encoders and their exponential-moving-average (EMA) teachers process each modality independently. We construct stop-gradient targets by translating each teacher token with the mean teacher representation of the paired modality. Consequently, each student predicts tokens from the unobserved modality, shifted by a summary of the modality it observes. Since this summary is predictable from the student's input, minimizing the prediction loss encourages both encoders to represent shared semantic content in a consistent coordinate system. Rather than enforcing alignment through an explicit constraint, U-JEPA allows it to emerge through non-contrastive objective. Crucially, this alignment is retained in the encoders themselves: after discarding the predictors, their representations remain non-collapsed and directly comparable across modalities.

Remarkably, U-JEPA achieves strong zero-shot cross-modal retrieval compared to contrastive models explicitly optimized for this task, despite never directly optimizing ranking objective. Under matched image--text pretraining conditions, U-JEPA improves zero-shot retrieval on COCO and Flickr30k benchmarks by up to 16\% relative over CLIP, while also achieving strong performance on zero-shot classification, vision benchmarks and composition understanding.  It further remains competitive with OpenCLIP\cite{ilharco2021openclip} and PowerCLIP\cite{kawamura2026powerclip}, which employs a stronger contrastive training recipe on retrieval tasks while outperforming them across most other benchmarks. Beyond global alignment, its token-level representations enable fine-grained retrieval through late interaction, while a substantially reduced modality gap indicates stronger cross-modal overlap in the representation space. Together, these findings demonstrate that pair-specific latent prediction can produce directly comparable multimodal representations without negative samples, cross-modal fusion modules, or explicit pairwise ranking.

Our main contributions are:
\begin{itemize}
    \item \textbf{Multimodal Alignment Through Latent Prediction.} We introduce U-JEPA, a non-contrastive framework that constructs pair-specific joint latent targets by translating modality-specific teacher representations with information from their paired modality, eliminating the need for learned joint teachers or cross-modal fusion modules.

    \item \textbf{Aligned Representations in modality specific encoders.} We demonstrate that predicting pair-specific latent targets with a shared predictor can induce alignment in independent modality encoders, enabling predictor-free cosine similarity comparisons and token-level late interaction at inference.

    \item \textbf{Strong Zero-Shot Generalization.} Under matched pretraining conditions, U-JEPA outperforms most contrastive baselines across zero-shot classification, retrieval, and compositional understanding, making it a non-contrastive alternative for multimodal training.
\end{itemize}

\begin{figure*}[t]
    \centering
    \includegraphics[width=0.95\textwidth]{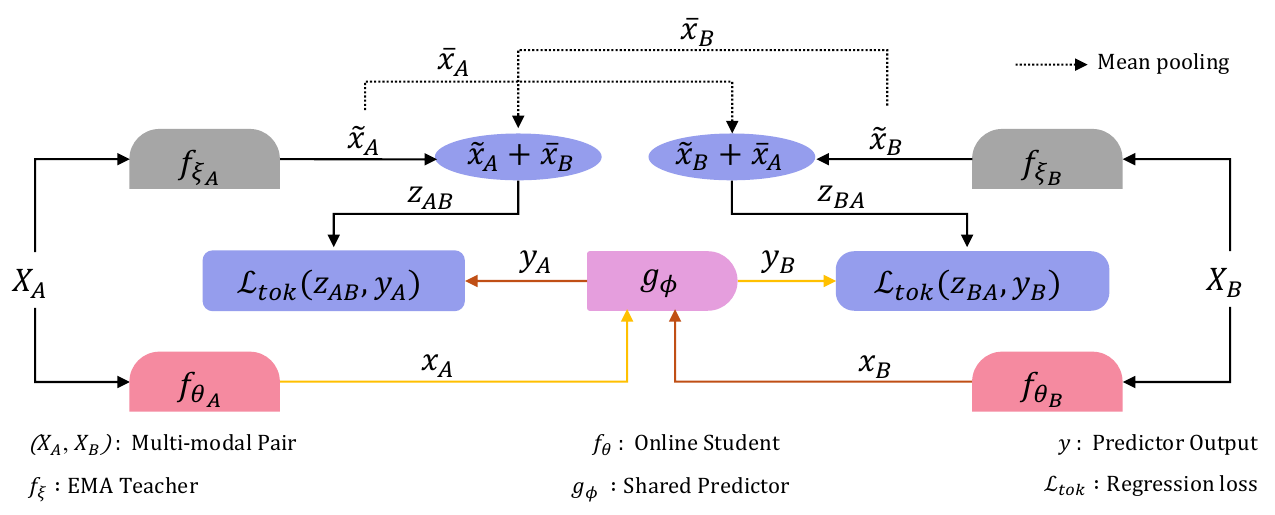}
    \caption{
\textbf{U-JEPA latent prediction objective}
Given a paired observation $(X_A,X_B)$, modality-specific students
$f_{\theta_A}$ and $f_{\theta_B}$ produce contextual token sequences, while
their EMA teachers $f_{\xi_A}$ and $f_{\xi_B}$ produce stop-gradient target
tokens $\tilde{x}_A$ and $\tilde{x}_B$. The valid teacher tokens are
mean-pooled to obtain $\bar{x}_A$ and $\bar{x}_B$ (dotted lines). Each mean is
injected into every target token of the opposite modality, yielding
$z_{AB,i}=(\tilde{x}_{A,i}+\bar{x}_B)$ and
$z_{BA,j}=(\tilde{x}_{B,j}+\bar{x}_A)$. A shared predictor $g_\phi$
maps the student context from $A$ to selected targets $z_{BA}$ (yellow) and
the context from $B$ to selected targets $z_{AB}$ (brown). The smooth-$L_1$
token loss $\mathcal{L}_{\mathrm{tok}}$ updates only the students and the
predictor. The local teacher tokens preserve target variation, while the
injected mean supplies a global summary of the observed source modality.
}
    \label{fig:objective}
\end{figure*}

\section{Related Works}

\subsection{Contrastive and Generative Methods}

Contrastive image--text pretraining explicitly aligns representations
by distinguishing matched from mismatched pairs, as exemplified by
CLIP \cite{radford2021clip} and ALIGN \cite{jia2021align}.
Subsequent work augments this formulation with cross-modal fusion
in ALBEF \cite{li2021align}, sigmoid-based pairwise objectives
in SigLIP \cite{zhai2023siglip}, masked modeling in FLAVA
\cite{singh2022flava}, and cross-modal self-distillation in
COSMOS \cite{kim2025cosmos}. Related approaches extend pairwise
alignment to audio--language
\cite{guzhov2022audioclip,elizalde2023clap}
and broader multimodal settings through image-anchored
\cite{girdhar2023imagebind} or language-anchored representations
\cite{zhu2024languagebind}. Despite differences in architecture
and auxiliary objectives, these methods retain explicit
cross-modal discrimination as a training signal. 

Generative pretraining learns cross-modal correspondences through
conditional token prediction. SimVLM \cite{wang2022simvlm}
adopts prefix language modeling, OFA \cite{wang2022ofa} uses
a sequence-to-sequence formulation, and CapPa
\cite{tschannen2023image} combines autoregressive and parallel
caption prediction. CoCa \cite{yu2022coca} and BLIP
\cite{li2022blip} jointly optimize generative and contrastive
objectives, while BEiT-3 \cite{wang2023beit3} applies masked
modeling to visual and textual tokens. In contrast, U-JEPA
operates through cross-modal latent prediction, without
contrastive supervision, explicit token generation, or
input reconstruction.

\subsection{Non-Contrastive Methods}

Non-contrastive self-supervision learns without negative samples, through
self-distillation, architectural asymmetry, and feature regularization
\cite{grill2020bootstrap,chen2021exploring,bardes2022vicreg}. data2vec
\cite{baevski2022data2vec} extends latent self-distillation to speech, vision, and
language, but predicts within each modality rather than across them.

In vision--language pretraining, nCLIP \cite{zhou2023noncontrastive} trails contrastive
CLIP on zero-shot classification and retrieval, motivating the hybrid xCLIP; SiamLIP,
BYOLIP, and BarLIP \cite{fini2023improved} add non-contrastive terms to contrastive
supervision for limited gains. Competitive cross-modal representations trained
from scratch, without negative pairs or fixed targets, thus remain an open problem.

\subsection{Joint-Embedding Predictive Architectures}

Joint-Embedding Predictive Architectures \cite{lecun2022path} predict latent targets
rather than reconstructing inputs or contrasting examples: I-JEPA \cite{assran2023ijepa}
for images, V-JEPA and its successors for spatiotemporal data
\cite{bardes2024vjepa,assran2025vjepa2,murlabadia2026vjepa21}. Across modalities,
M3-JEPA \cite{lei2025m3jepa} uses a mixture-of-experts predictor with contrastive and
regularization terms, VL-JEPA \cite{chen2026vljepa} predicts continuous text embeddings
from visual inputs, and TC-JEPA \cite{huang2026tcjepa} conditions masked visual
prediction on text through sparse cross-attention. Closest to our setting, LeVLJEPA
\cite{kuhn2026levljepa} pretrains end-to-end and bidirectionally with modality-specific
predictors, stop-gradient targets, and per-modality distributional regularization; its
dense visual features are strong, but its zero-shot classification and global alignment
trail contrastive baselines. Beyond vision--language, AV-JEPA \cite{robson2026avjepa}
uses early fusion and modality dropout, X-JEPA \cite{choudhury2026xjepa} explicit
prediction-space alignment, and CR-JEPA \cite{hossain2026crjepa} a shared backbone with
retrieval heads. NOVA
\cite{kuhn2026noncontrastive} removes contrastive supervision through predictive
alignment and Sketched Isotropic Gaussian Regularization \cite{balestriero2025lejepa},
but predicts into a frozen domain-specific text encoder, so its semantic targets are
pretrained rather than jointly learned. 

These methods establish that joint representation learning is feasible without negative
pairs, but each supplies cross-modal correspondence by a different
mechanism---contrastive terms, text conditioning, modality-specific predictors, fusion,
or auxiliary alignment etc. U-JEPA attempts to learn alignment that persists within the encoder outputs when training only components are discarded, preserving the dual encoder efficiency at inference.

\begin{table*}[t]
\centering
\caption{\textbf{Zero-shot image--text retrieval.} Recall@K (\%) on MS-COCO
and Flickr30K with ViT-B/16 image encoders. The public PowerCLIP, OpenCLIP, LeVLJEPA and U-JEPA are evaluated with same pipeline. The prior
non-contrastive methods, nCLIP and LeVLJEPA, are shown for reference.}
\label{tab:retrieval}
\small
\setlength{\tabcolsep}{3pt}
\renewcommand{\arraystretch}{1.12}
\resizebox{\textwidth}{!}{%
\begin{tabular}{@{}llrrrrrrrrrrrr@{}}
\toprule
& & \multicolumn{6}{c}{\textbf{MS-COCO}~\cite{lin2014coco}}
& \multicolumn{6}{c}{\textbf{Flickr30K}~\cite{young2014flickr}} \\
\cmidrule(lr){3-8}\cmidrule(lr){9-14}
& & \multicolumn{3}{c}{Image $\rightarrow$ Text}
& \multicolumn{3}{c}{Text $\rightarrow$ Image}
& \multicolumn{3}{c}{Image $\rightarrow$ Text}
& \multicolumn{3}{c}{Text $\rightarrow$ Image} \\
\cmidrule(lr){3-5}\cmidrule(lr){6-8}\cmidrule(lr){9-11}\cmidrule(lr){12-14}
\textbf{Method} & \textbf{Dataset used}
& R@1 & R@5 & R@10 & R@1 & R@5 & R@10
& R@1 & R@5 & R@10 & R@1 & R@5 & R@10 \\
\midrule
\multicolumn{14}{l}{\textit{Contrastive baselines}} \\
CLIP~\cite{radford2021clip} & CC12M & 32.76 & 60.18 & 71.08 & 22.59 & 46.57 & 58.20 & 58.70 & 84.70 & 91.90 & 45.68 & 71.90 & 80.56 \\
PowerCLIP-S~\cite{kawamura2026powerclip} & CC12M & 37.30 & \textbf{64.90} & \textbf{75.60} & \textbf{27.00} & \textbf{52.90} & \textbf{64.00} & 62.40 & \textbf{88.50} & \textbf{94.20} & 50.40 & 76.60 & 84.60 \\
OpenCLIP~\cite{ilharco2021openclip,cherti2023reproducible} & CC12M & 36.24 & 62.88 & 74.32 & 26.86 & 51.72 & 63.21 & 62.90 & 87.70 & 93.40 & 50.14 & 76.70 & 84.38 \\
\midrule
\multicolumn{14}{l}{\textit{Non-contrastive methods}} \\
nCLIP~\cite{zhou2023noncontrastive} & IT35M & 21.40 & 47.90 & 61.60 & 13.40 & 37.10 & 50.30 & 26.20 & 53.30 & 67.90 & 20.10 & 48.50 & 61.10 \\
LeVLJEPA$^{*}$~\cite{kuhn2026levljepa} & DataComp-L & 17.18 & 36.64 & 48.24 & 11.24 & 27.33 & 37.79 & 30.60 & 58.40 & 69.00 & 22.60 & 49.50 & 61.58 \\
U-JEPA (mean) & CC12M & 37.56 & 62.50 & 72.34 & 23.89 & 46.54 & 58.18 & 65.70 & 85.70 & 92.70 & 47.28 & 75.36 & 83.12 \\
\midrule
\multicolumn{14}{l}{\textit{Token-level late interaction with query-side MaxSim}} \\
PowerCLIP-S & CC12M & 28.36 & 53.76 & 65.98 & 16.65 & 38.95 & 51.39 & 56.10 & 82.40 & 89.60 & 37.86 & 67.60 & 77.76 \\
OpenCLIP  & CC12M & 5.48 & 15.72 & 22.96 & 5.66 & 15.74 & 23.05 & 9.80 & 23.30 & 33.40 & 11.18 & 26.80 & 35.04 \\
LeVLJEPA$^\ast$  & DataComp-L & 5.32 & 13.76 & 20.72 & 5.21 & 14.49 & 21.76 & 11.10 & 25.90 & 36.00 & 8.30 & 21.12 & 29.28 \\
\textbf{U-JEPA } & CC12M & \textbf{38.14} & 63.60 & 73.38 & 26.24 & 49.68 & 61.31 & \textbf{67.50} & 86.50 & 91.80 & \textbf{51.00} & \textbf{77.58} & \textbf{84.94} \\
\bottomrule
\end{tabular}}
\vspace{2pt}
\parbox{\textwidth}{\footnotesize
$^{*}$LeVLJEPA uses its cross-modal predictors at inference, without predictors, zero-shot retrieval is at near chance}
\end{table*}

\section{Method}
\label{sec:method}

U-JEPA trains independent modality encoders by predicting EMA-teacher tokens
augmented with the paired modality's mean representation. A shared predictor
performs the two regression tasks. The Fig.\ref{fig:objective} illustrates the token-prediction pathway.

\subsection{Student--Teacher Encoders}
\label{sec:method_encoders}

For each modality $m\in\{A,B\}$, an online encoder $f_{\theta_m}$ and an
EMA teacher $f_{\xi_m}$ process the input independently, with
\begin{equation}
    \xi_m\leftarrow\tau\xi_m+(1-\tau)\theta_m.
    \label{eq:ema_update}
\end{equation}
The students produce $x_A$ and $x_B$; the teachers produce token sequences
$\tilde{x}_A=\{\tilde{x}_{A,i}\}_{i=1}^{N}$ and
$\tilde{x}_B=\{\tilde{x}_{B,j}\}_{j=1}^{M}$. All encoders process complete inputs, and context
selection occurs after student encoding. The towers share no parameters or
cross-modal attention; padding is excluded from pooling and losses.

\subsection{Cross-Modal Target Injection}
\label{sec:method_injection}

The valid-token teacher means are
\begin{equation}
    \bar{x}_A=\frac{1}{N}\sum_{i=1}^{N}\tilde{x}_{A,i},\qquad
    \bar{x}_B=\frac{1}{M}\sum_{j=1}^{M}\tilde{x}_{B,j}.
    \label{eq:pooled_means}
\end{equation}
Each teacher token is shifted by the mean of the opposite modality:
\begin{equation}
    z_{AB,i}=\tilde{x}_{A,i}+\alpha\bar{x}_B,\qquad
    z_{BA,j}=\tilde{x}_{B,j}+\alpha\bar{x}_A.
    \label{eq:injected_tokens}
\end{equation}
with $\alpha=1$. The first subscript identifies the target-token modality;
the second identifies the injected modality. 

\subsection{Symmetric Token Prediction}
\label{sec:method_prediction}

The Transformer predictor $g_\phi$ is shared across both directions. From
selected student contexts $C_A\subset x_A$ and $C_B\subset x_B$ and
position-specific queries, it predicts
\begin{equation}
\begin{aligned}
    y_{B,j}&=g_\phi(C_A,q_j^B), &&j\in\mathcal{T}_B,\\
    y_{A,i}&=g_\phi(C_B,q_i^A), &&i\in\mathcal{T}_A.
\end{aligned}
    \label{eq:cross_modal_predictions}
\end{equation}
Accordingly, $y_B$ is regressed against $z_{BA}$ and $y_A$ against $z_{AB}$,
as labeled in the figure. The token regression objective is
\begin{equation}
\begin{aligned}
\mathcal{L}_{\mathrm{tok}}=\frac12\bigg[&
\frac{1}{|\mathcal{T}_B|}\sum_{j\in\mathcal{T}_B}
\mathcal{D}(y_{B,j},\operatorname{sg}[z_{BA,j}])\\
&+\frac{1}{|\mathcal{T}_A|}\sum_{i\in\mathcal{T}_A}
\mathcal{D}(y_{A,i},\operatorname{sg}[z_{AB,i}])\bigg],
\end{aligned}
\label{eq:token_prediction_loss}
\end{equation}
where $\mathcal{D}$ is feature-averaged smooth-$L_1$ and
$\operatorname{sg}$ denotes stop-gradient. Each direction is averaged
over its requested tokens, and the two directions contribute equally. For image-to-text prediction, 49 of 196 encoded image patches
form the context and four text spans of four positions are queried. For
text-to-image prediction, up to 77 encoded text tokens form the context and
four image blocks of 30 patches are queried. Text padded positions are excluded from loss calculation. 

\paragraph{What exactly is predicted.}
\label{sec:method_task}
Regressing only pooled representations would replace each target sequence by its mean
and discard the variation among its tokens; in our experiments, this degrades
cross-modal alignment. We
instead retain the individual teacher tokens and inject the observed modality's mean:
the image context predicts $z_{BA,j}=(\tilde{x}_{B,j}+\bar{x}_A)$ at text position $j$, and
the text context predicts $z_{AB,i}=(\tilde{x}_{A,i}+\bar{x}_B)$ at image position $i$. Thus,
the paired mean provides the cross-modal offset while the local token preserves the
target sequence's variation. The predictor regresses these contextual teacher latents.

Following the target-sampling convention of I-JEPA~\cite{assran2023ijepa}, we request
contiguous regions rather than every target token: four spans of four text positions and
four image blocks of 30 patches. This structured subsampling retains local,
position-specific targets while reducing the number of predictor queries and hence the
training cost.

Although the requested latents are position-specific, they are not uniquely determined
by the source modality: an image admits multiple captions, and a caption does not specify
all local image content. A deterministic predictor trained with smooth-$L_1$ therefore
approaches a conditional mean of the target distribution. This gives a
well-defined regression solution despite the one-to-many mapping, but it can average
over source-ambiguous variation and thereby weaken the identity of individual target
tokens, which we analyze in Sec \ref{sec:results}

\paragraph{Motivation and scope.}
Consider one direction of Eq.~\ref{eq:cross_modal_predictions}: given context $C_A$, the student must predict $\tilde{x}_{B,i} + \alpha\bar{x}_A$, a token of $X_B$ shifted by the mean of $X_A$. If the target were purely $\tilde{x}_{B,i}$, the predictor could solve the task by acting as a standalone translator from $X_A$ to $X_B$ without the two encoders ever sharing a coordinate system. Instead, because the target is a function of both modalities, the predictor's output must simultaneously match statistics of $X_A$ and $X_B$ in a single latent space---forcing it to preserve $X_A$'s representation while predicting the displacement onto $\tilde{x}_{B,i}$.

Predictability depends on the statistical dependence between the modalities: absent shared information, $\tilde{x}_{B,i}$ is an arbitrary function of unobserved content, and the risk is irreducible. Where semantic content is shared, coupling both modalities in the target pressures the encoders to represent that content in a common coordinate system: the closer the encoders' coordinates, the smaller the residual translation the predictor must supply on top of $\alpha\bar{x}_A$, reducing to a local offset in the limit. Encoder alignment thus emerges as a plausible solution to an objective that never explicitly enforces it.

This argument is motivational rather than a formal guarantee, as two failure modes remain: representation collapse---where a constant target renders every displacement predictable---and predictor absorption, where a sufficiently expressive predictor learns an invertible reparameterization between misaligned encoders. As we show empirically, generating the latent targets via an EMA teacher is sufficient to prevent the student from trivially pulling the target toward its prediction and to suppress collapse; the remaining design choices are evaluated in Sec \ref{sec:ablation}.

\subsection{Auxiliary Pair Distillation}
\label{sec:method_summary}
Latent prediction with mixed targets produces strong alignment on its own
(Sec.~\ref{sec:ablation}), but it does not rule out the predictor absorbing
part of the cross-modal mapping. We hypothesize that a regularizer applied
outside the predictor path discourages this absorption, so that more of the
alignment resides in the encoders. We introduce such a term, auxiliary pair
distillation.

The objective is independent of $g_\phi$ and acts on the student and teacher
\texttt{[CLS]} states $s_m^\theta$ and $s_m^\xi$. Using
$\bar{x}_m^{\mathrm{raw}}$ for the teacher-token mean before LayerNorm,
the pair-specific teacher summaries are
\begin{equation}
\widetilde{s}_A^\xi=s_A^\xi+\bar{x}_B^{\mathrm{raw}},
\qquad
\widetilde{s}_B^\xi=s_B^\xi+\bar{x}_A^{\mathrm{raw}}.
\label{eq:summary_injection}
\end{equation}
A shared prototype head $h_\theta$ and its EMA copy $h_\xi$ yield
\begin{equation}
\begin{aligned}
    p_m&=\operatorname{softmax}(h_\theta(s_m^\theta)/T_s),\\
    q_m&=\operatorname{softmax}([h_\xi(\widetilde{s}_m^\xi)-c]/T_t),
\end{aligned}
    \label{eq:summary_assignments}
\end{equation}
where $c$ is an EMA center of teacher logits, $K=1024$ is the number of
prototypes, $T_s=0.1$, $T_t=0.04$, and the center momentum is $0.9$. The
pair distillation loss and the complete objective are
\begin{equation}
    \mathcal{L}_{\mathrm{pair}}
    =\frac{H(q_B,p_A)+H(q_A,p_B)}{2},
    \label{eq:summary_loss}
\end{equation}
\begin{equation}
    \mathcal{L}_{\mathrm{U\text{-}JEPA}}
    =\mathcal{L}_{\mathrm{tok}}+0.1\,\mathcal{L}_{\mathrm{pair}}.
    \label{eq:total_loss}
\end{equation}
The prototype head, centering, and temperatures follow
DINO~\cite{caron2021emerging}, but the objective differs. DINO matches two
views of an input to the same output, whereas here each student summary is
matched to its own pair-specific target. The image student is matched to
$q_B$, built from the text teacher's summary and the image teacher's token
mean; the text student is matched to $q_A$, built symmetrically. The two
targets differ, so the loss does not constrain the two summaries to coincide,
and each student can match the other modality's component only through
content the pair shares. The effect of this term is evaluated in
Sec.~\ref{sec:ablation}.

\subsection{Compute Budget}
\label{sec:method_compute}

Both towers are identical to the contrastive baseline, so
the cost of the method is exactly the teacher pass and the predictor, and both are
discarded after pretraining. Table~\ref{tab:compute} states the budget per
image--caption pair against a CLIP baseline at that same architecture. Pretraining
costs $1.75\times$ per pair---$19\%$ teacher, $23\%$ predictor---so our $60$ epochs are
the compute of $105$ contrastive ones. we highlight that longer pretraining does not benefit contrastive baselines and they saturate much earlier than 60 epochs and retrieval degrading in later epochs.

\begin{table}[t]
\centering
\caption{\textbf{training budget, per image--caption pair.}
U-JEPA Base against a CLIP baseline with the same two towers, at $224{\times}224$
($N{=}196$) and $T{=}77$}
\label{tab:compute}
\small
\setlength{\tabcolsep}{5pt}
\renewcommand{\arraystretch}{1.04}
\begin{tabular*}{\columnwidth}{@{\extracolsep{\fill}}lrrr@{}}
\toprule
 & \textbf{U-JEPA} & \textbf{CLIP} & \textbf{Ratio} \\
\midrule
Trainable parameters        & 163\,M & 149\,M & $1.09\times$ \\
Pretraining FLOPs           & 215\,G & 123\,G & $1.75\times$ \\
Inference parameters        & 149\,M & 149\,M & $1.00\times$ \\
Inference FLOPs             & 41\,G  & 41\,G  & $1.00\times$ \\
\bottomrule
\end{tabular*}
\end{table}

\section{Experimental Setup}

We pretrain U-JEPA from scratch on CC3M (ViT-S and ViT-B; batch size 1,024)
and CC12M (ViT-B; batch size 4,096), with a 12-layer, width-512 text encoder.
We use AdamW with a learning rate warmed up linearly from \(2\times10^{-4}\)
to \(5\times10^{-4}\) over five epochs and then held constant, cosine weight decay
from \(0.04\) to \(0.4\), teacher momentum rising linearly from \(0.996\) to
\(1.0\). For U-JEPA inference, EMA teachers are used as final model weights.

CLIP trained for 105 epochs, matching U-JEPA's compute, saturates early with
no consistent later gain, consistent with \citet{Li2024ScalingDownCLIP}. We also
evaluate to our knowledge, the strongest public OpenCLIP checkpoint trained on CC12M with
original captions; it uses a larger batch size (8,192) and
a more tuned recipe. U-JEPA uses bidirectional text attention; in our CLIP
runs, switching from causal to bidirectional attention did not help.
LeVLJEPA is the public checkpoint pretrained on DataComp-L (92M pairs).
\section{Results and Discussion}
\label{sec:results}

U-JEPA learns transferable and aligned image--text representations without contrastive
pretraining. Mean pooling and MaxSim measure global and token-level
alignment directly from encoder outputs, without learned inference parameters.
The predictor and online encoders are discarded after pretraining.
Reported MaxSim scores use query-side MaxSim operation. Unless otherwise stated all results use encoder-only readouts by discarding training only components.
\begin{table}[t]
\centering
\caption{\textbf{CC3M zero-shot retrieval.} R@1/R@5 (\%) on COCO and Flickr30K.}
\label{tab:retrieval_cc3m}
\small
\setlength{\tabcolsep}{2.5pt}
\renewcommand{\arraystretch}{1.12}
\resizebox{\columnwidth}{!}{%
\begin{tabular}{@{}lrrrr@{}}
\toprule
& \multicolumn{2}{c}{COCO} & \multicolumn{2}{c}{Flickr30K} \\
\cmidrule(lr){2-3}\cmidrule(lr){4-5}
Method & I$\rightarrow$T & T$\rightarrow$I & I$\rightarrow$T & T$\rightarrow$I \\
\midrule
\multicolumn{5}{l}{\textit{ViT-S/16}} \\
CLIP & 17.30 / 39.00 & \textbf{13.10} / \textbf{31.20} & 35.20 / \textbf{62.30} & 25.40 / 49.12 \\
U-JEPA (mean) & \textbf{19.54} / \textbf{40.22} & 12.73 / 30.41 & \textbf{35.70} / 62.00 & 25.30 / 52.58 \\
U-JEPA (MaxSim) & 18.86 / 39.94 & 12.87 / 30.77 & 34.20 / 61.20 & \textbf{25.56} / \textbf{53.14} \\
\midrule
\multicolumn{5}{l}{\textit{ViT-B/16}} \\
CLIP & 18.60 / 41.10 & 13.90 / 32.80 & 36.10 / 65.10 & 26.30 / 52.40 \\
U-JEPA (mean) & \textbf{20.60} / 41.84 & 12.77 / 30.56 & 38.20 / 65.60 & 27.44 / 53.92 \\
U-JEPA (MaxSim) & 19.84 / \textbf{42.62} & \textbf{14.32} / \textbf{32.95} & \textbf{39.30} / \textbf{66.50} & \textbf{30.24} / \textbf{56.12} \\
\bottomrule
\end{tabular}}
\end{table}

\subsection{Zero-Shot Image--Text Retrieval}
\label{sec:results_retrieval}

U-JEPA improves zero-shot retrieval over the compared both contrastive and non-contrastive methods
(Tab.~\ref{tab:retrieval}). Mean pooling exceeds CLIP baseline in both directions on
COCO and Flickr30K, establishing direct global alignment.

Against the stronger OpenCLIP and PowerCLIP-S, MaxSim achieves higher bidirectional
R@1 sums on both datasets, although they leads at some higher-recall
cutoffs. U-JEPA gains over mean pooling reveals matching information beyond a single global embedding.

Unlike U-JEPA, LeVLJEPA requires cross-modal predictors for its reported
retrieval; its pre-predictor embeddings yield near-chance performance.
Table~\ref{tab:retrieval_cc3m} compares both encoder scales on CC3M.
The query-side MaxSim results reveal substantial degradation in zero-shot retrieval performance across the evaluated methods, whereas U-JEPA benefits from it.

\subsection{Zero-Shot Classification}
\label{sec:results_classification}

U-JEPA transfers from instance matching to objects, scenes, and textures
(Tab.~\ref{tab:zeroshot_classification}). A single frozen mean readout leads
most evaluated benchmarks, combining direct retrieval alignment with
class-level transfer. These results assess whether representations learned
from paired instances also organize categories that were never explicitly
used as training labels.

\begin{table*}[t]
\centering
\caption{\textbf{Classification and ImageNet robustness.} Accuracy (\%) with ViT-B/16 image encoders pretrained on CC12M. Pets, Flowers, and Caltech-101 use mean-per-class accuracy in our evaluations; other datasets use top-1. Our linear probes average three seeds. Flwr: Flowers-102; Cal: Caltech-101; Euro: EuroSAT; RESISC: RES; Sk.: ImageNet-Sketch.}
\label{tab:zeroshot_classification}
\label{tab:imagenet}
\small
\setlength{\tabcolsep}{2.5pt}
\renewcommand{\arraystretch}{1.12}
\resizebox{\textwidth}{!}{%
\begin{tabular}{@{}lrrrrrrrrrrrrrrrrrr@{}}
\toprule
& \multicolumn{11}{c}{\textbf{Zero-shot classification}}
& \multicolumn{6}{c}{\textbf{ImageNet zero-shot}} & \textbf{LP} \\
\cmidrule(lr){2-12}\cmidrule(lr){13-18}\cmidrule(lr){19-19}
\textbf{Method}
 & \shortstack{Food\\\cite{bossard2014food101}} & \shortstack{C10\\\cite{krizhevsky2009tiny}} & \shortstack{C100\\\cite{krizhevsky2009tiny}} & \shortstack{SUN\\\cite{xiao2010sun}} & \shortstack{Cars\\\cite{krause2013cars}} & \shortstack{DTD\\\cite{cimpoi2014textures}} & \shortstack{Pets\\\cite{parkhi2012pets}} & \shortstack{Flwr\\\cite{nilsback2008flowers}} & \shortstack{Cal\\\cite{feifei2004learning}} & \shortstack{Euro\\\cite{helber2019eurosat}} & \shortstack{RES\\\cite{cheng2017remote}} & \shortstack{1K\\\cite{deng2009imagenet}} & \shortstack{V2\\\cite{recht2019imagenet}} & \shortstack{A\\\cite{hendrycks2021natural}} & \shortstack{R\\\cite{hendrycks2021faces}} & \shortstack{O\\\cite{hendrycks2021natural}} & \shortstack{Sk.\\\cite{wang2019sketch}} & 1K \\
\midrule
\multicolumn{19}{l}{\textit{Contrastive baselines}} \\
CLIP~\cite{radford2021clip} & 42.3 & 57.7 & 25.0 & 44.1 & 17.0 & 16.5 & 53.9 & 26.0 & 73.5 & 18.7 & 26.5 & 36.1 & 30.7 & 8.0 & 47.6 & 38.4 & 24.9 & 62.3 \\
PowerCLIP-S~\cite{kawamura2026powerclip} & 50.3 & 74.7 & 43.5 & 48.7 & 22.9 & 21.5 & 58.7 & \textbf{32.4} & 75.7 & 30.8 & \textbf{37.5} & 40.3 & 34.8 & 11.2 & 53.2 & 40.2 & 28.7 & 65.8 \\
OpenCLIP~\cite{ilharco2021openclip,cherti2023reproducible} & 48.8 & 66.2 & 38.8 & 47.6 & \textbf{26.3} & 19.6 & 62.5 & 32.0 & 75.1 & 25.7 & 28.3 & 42.0 & 36.6 & 10.5 & 53.9 & \textbf{44.0} & 30.1 & 67.3 \\
\midrule
\multicolumn{19}{l}{\textit{Non-contrastive}} \\
\textbf{U-JEPA (mean)} & \textbf{53.2} & \textbf{77.6} & \textbf{52.7} & \textbf{51.2} & 24.5 & \textbf{24.6} & \textbf{69.1} & 31.3 & \textbf{79.1} & 32.8 & 33.8 & \textbf{44.3} & \textbf{38.1} & \textbf{13.4} & \textbf{66.6} & 39.5 & \textbf{36.2} & \textbf{70.2} \\
\bottomrule
\end{tabular}}
\end{table*}

\subsection{ImageNet Classification and Distribution Shifts}
\label{sec:results_imagenet}

U-JEPA's ImageNet advantage over OpenCLIP persists under linear probing
(Tab.~\ref{tab:imagenet}), demonstrating discriminative image features.
Transfer remains strong on ImageNet-V2, -A, -R, and Sketch without adaptation;
OpenCLIP leads on ImageNet-O. The agreement between zero-shot recognition
and linear probing supports an advantage in the visual representation,
beyond compatibility with text-derived class prototypes.

\subsection{Compositional Understanding}
\label{sec:results_composition}
\label{sec:results_sugarcrepe}
\label{sec:results_winoground}

Tables~\ref{tab:sugarcrepe} and~\ref{tab:composition_extended} show fine-grained
semantic discrimination across caption edits, attribute binding, and linguistic
foils. Gains on SugarCrepe, ARO attribution, and VALSE actions and counting
extend transfer beyond global recognition. MaxSim accesses additional token-level
distinctions despite the representations' strong global component.

This profile accords with U-JEPA objective: shared scene content
constrains prediction, whereas images do not uniquely determine word order.
Regression across admissible descriptions may therefore preserve semantics
more reliably than ordering. Weaker ARO order results support this
interpretation without establishing causality; Winoground also exposes
persistent difficulty with paired relational assignment. 

\begin{table}[t]
\centering
\caption{\textbf{SugarCrepe compositionality.} Accuracy (\%). Obj/Att/Rel average subsets within each category; Overall weights all 7,511 examples equally.}
\label{tab:sugarcrepe}
\small
\setlength{\tabcolsep}{3pt}
\renewcommand{\arraystretch}{1.12}
\begin{tabular}{@{}llrrrr@{}}
\toprule
\textbf{Method} & \textbf{Dataset used} & Obj & Att & Rel & Overall \\
\midrule
CLIP~\cite{radford2021clip} & CC12M & 74.3 & 71.5 & \textbf{67.6} & 75.0 \\
OpenCLIP~\cite{ilharco2021openclip,cherti2023reproducible} & CC12M & 76.0 & \textbf{71.8} & 65.9 & 75.4 \\
LeVLJEPA~\cite{kuhn2026levljepa} & DataComp-L & 52.1 & 50.3 & 52.3 & 52.1 \\
U-JEPA (mean) & CC12M & 75.6 & 68.1 & 61.0 & 74.4 \\
\midrule
\textbf{U-JEPA (MaxSim)} & CC12M & \textbf{79.0} & 71.0 & 66.5 & \textbf{77.9} \\
\bottomrule
\end{tabular}
\end{table}

\begin{table}[t]
\centering
\caption{\textbf{Compositional understanding.} Frozen-encoder accuracy (\%). All models use CC12M except LeVLJEPA (DataComp-L), evaluated without predictors. VALSE averages 11 pieces.}
\label{tab:winoground}
\label{tab:composition_extended}
\small
\setlength{\tabcolsep}{1.5pt}
\resizebox{\columnwidth}{!}{%
\begin{tabular}{@{}lrrrrr@{}}
\toprule
& \multicolumn{3}{c}{Baselines} & \multicolumn{2}{c}{U-JEPA} \\
\cmidrule(lr){2-4}\cmidrule(lr){5-6}
Benchmark & CLIP & OpenCLIP & LeVLJEPA & Mean & MaxSim \\
\midrule
\multicolumn{6}{l}{\textbf{Winoground}~\cite{thrush2022winoground}} \\
\quad Text & 22.5 & 23.0 & 18.0 & 24.5 & \textbf{27.3} \\
\quad Image & 9.3 & \textbf{11.5} & 9.8 & 9.0 & 9.5 \\
\quad Group & 6.3 & 6.8 & 3.5 & 7.3 & \textbf{7.5} \\
\midrule
\multicolumn{6}{l}{\textbf{ARO}~\cite{yuksekgonul2023when}} \\
\quad Attribute & 60.0 & 59.6 & 50.5 & 63.7 & \textbf{67.8} \\
\quad Relation & 51.4 & 46.7 & 51.3 & 51.4 & \textbf{53.9} \\
\quad COCO order & 28.2 & \textbf{31.2} & 16.1 & 18.8 & 21.1 \\
\quad Flickr order & 32.8 & \textbf{35.6} & 16.5 & 25.0 & 27.8 \\
\midrule
VALSE~\cite{parcalabescu2022valse} & 59.1 & 58.6 & 50.4 & 62.3 & \textbf{65.3} \\
\bottomrule
\end{tabular}}
\end{table}

\subsection{Dense Semantic Segmentation}
\label{sec:results_dense}

To test whether global coherence compromises spatial discrimination, we fit
linear classifiers to frozen ADE20K and COCO-Stuff patch features
(Tab.~\ref{tab:dense_segmentation}). Under identical probe protocols, U-JEPA
leads both references on both metrics: strong global representations retain
linearly accessible spatial semantics.

\begin{table}[t]
\centering
\caption{\textbf{Frozen dense segmentation.} Validation mIoU and mean per-class pixel accuracy (mAcc), in \%, averaged over three seeds. All models use ViT-B/16 backbones, 224-pixel inputs, and a single linear head on 768-dimensional pre-projection patch features. Training and evaluation protocols are matched across models.}
\label{tab:dense_segmentation}
\small
\setlength{\tabcolsep}{3pt}
\renewcommand{\arraystretch}{1.12}
\resizebox{\columnwidth}{!}{%
\begin{tabular}{@{}llrrrr@{}}
\toprule
& & \multicolumn{2}{c}{ADE20K~\cite{zhou2017ade20k}} & \multicolumn{2}{c}{COCO-S~\cite{caesar2018cocostuff}} \\
\cmidrule(lr){3-4}\cmidrule(lr){5-6}
\textbf{Method} & \textbf{Pretraining} & mIoU & mAcc & mIoU & mAcc \\
\midrule
CLIP~\cite{radford2021clip} & CC12M & 27.11 & 33.86 & 29.61 & 39.73 \\
PowerCLIP-S~\cite{kawamura2026powerclip} & CC12M & 28.11 & 35.42 & 30.93 & 41.17 \\
OpenCLIP~\cite{ilharco2021openclip,cherti2023reproducible} & CC12M & 27.18 & 33.95 & 30.27 & 40.38 \\
LeVLJEPA~\cite{kuhn2026levljepa} & DataComp-L & 26.32 & 32.94 & 30.06 & 39.78 \\
\textbf{U-JEPA} & CC12M & \textbf{30.52} & \textbf{39.35} & \textbf{33.57} & \textbf{44.84} \\
\bottomrule
\end{tabular}}
\end{table}

\subsection{Dense-Caption Matching}
\label{sec:results_dci}

The official single-summary sDCI benchmark~\cite{urbanek2024picture}
separates region--caption matching from targeted negative discrimination
(Tab.~\ref{tab:dci}). U-JEPA leads matching with region crops, while OpenCLIP
leads on complete images. Mean pooling favors region matching; MaxSim improves
negative discrimination across All, Base, and Hardest, exposing complementary information
in the same frozen representations. Improved negative discrimination is
therefore distinct from improved matching, and neither readout uniformly
dominates across the benchmark.

\begin{table}[t]
\centering
\caption{\textbf{Zero-shot sDCI evaluation.} Accuracy (\%) under the authors'
ordered batch-eight protocol~\cite{urbanek2024picture}. All includes complete
images and crops (1,382 pairs); Base uses complete images (108); Hardest
changes only the selected negatives. M: matching; N: negative discrimination.
CLIP~\cite{radford2021clip}, OpenCLIP~\cite{ilharco2021openclip,cherti2023reproducible},
and U-JEPA use CC12M; LeVLJEPA~\cite{kuhn2026levljepa} uses DataComp-L and retains
its predictors.}
\label{tab:dci}
\small
\setlength{\tabcolsep}{1.5pt}
\begin{tabular*}{\columnwidth}{@{\extracolsep{\fill}}lrrrrrr@{}}
\toprule
& \multicolumn{2}{c}{All} & \multicolumn{2}{c}{Base} & \multicolumn{2}{c}{Hardest} \\
\cmidrule(lr){2-3}\cmidrule(lr){4-5}\cmidrule(lr){6-7}
Method & M & N & M & N & M & N \\
\midrule
CLIP & 36.7 & 63.8 & 94.4 & 74.1 & 36.7 & 57.2 \\
OpenCLIP & 37.4 & 61.8 & \textbf{99.1} & 78.7 & 37.4 & 56.2 \\
LeVLJEPA & 35.5 & 55.0 & 86.1 & 50.0 & 35.5 & 51.7 \\
U-JEPA (mean) & \textbf{39.9} & 63.0 & 96.3 & 75.0 & \textbf{39.9} & 57.2 \\
\midrule
U-JEPA (MaxSim) & 39.4 & \textbf{65.0} & 96.3 & \textbf{79.6} & 39.4 & \textbf{59.3} \\
\bottomrule
\end{tabular*}
\end{table}

\subsection{Mechanism Ablations}
\label{sec:ablation}

We ablate U-JEPA around the final shared predictor(width-512), $\alpha=1$ and bidirection text encoder configuration with both latent prediction and auxiliary loss. Every variant uses the same fixed 50\% subset of CC12M, training schedule, data
ordering, and evaluation protocol, and changes only the component named in
Table~\ref{tab:ablations}. We report the strongest mean-pooled target-encoder readout
on COCO 5K using raw cosine similarity. Sum R@1 is the sum of image-to-text and text-to-image R@1.

\begin{table}[t]
\centering
\caption{\textbf{Mechanism and architecture ablations.}
Raw COCO 5K retrieval (\%) on the fixed 50\% CC12M ablation setting.
$\Delta$ is the change in Sum R@1 relative to the U-JEPA configuration.
$^\dagger$ denotes the reference configuration. All results use mean-pooled token
representations.}
\label{tab:ablations}
\small
\setlength{\tabcolsep}{4pt}
\renewcommand{\arraystretch}{1.06}
\resizebox{\columnwidth}{!}{
\begin{tabular}{@{}lrrrr@{}}
\toprule
\textbf{Variant} & \textbf{I$\rightarrow$T R@1} & \textbf{T$\rightarrow$I R@1}
& \textbf{Sum} & \textbf{$\Delta$} \\
\midrule
\textbf{Full U-JEPA}$^\dagger$
& \textbf{36.06} & \textbf{20.95} & \textbf{57.01} & -- \\

\addlinespace[2pt]
\multicolumn{5}{@{}l}{\textit{Learning objective}} \\
\quad Latent prediction only & 30.74 & 18.02 & 48.76 & $-8.25$ \\
\quad Pair-distillation only & 2.20 & 2.25 & 4.45 & $-52.56$ \\
\quad Pair-distillation w/o cls token & 30.66 & 19.04 & 49.70 & $-7.31$ \\

\addlinespace[2pt]
\multicolumn{5}{@{}l}{\textit{Target construction}} \\
\quad No mixing ($\alpha{=}0$)
& 15.14 & 10.02 & 25.16 & $-31.85$ \\
\quad Weaker mixing ($\alpha{=}0.5$) & 27.42 & 17.39 & 44.81 & $-12.20$ \\
\quad Stronger mixing ($\alpha{=}1.5$) & 30.98 & 18.31 & 49.29 & $-7.72$ \\
\quad Pooled teacher mean (MLP) & 4.64 & 3.62 & 8.26 & $-48.75$ \\

\addlinespace[2pt]
\multicolumn{5}{@{}l}{\textit{Pair correspondence}} \\
\quad Random image--text pairs & 0.08 & 0.04 & 0.12 & $-56.89$ \\

\addlinespace[2pt]
\multicolumn{5}{@{}l}{\textit{Text encoder}} \\
\quad Causal/EOS & 22.76 & 12.89 & 35.65 & $-21.36$ \\

\addlinespace[2pt]
\multicolumn{5}{@{}l}{\textit{Predictor role}} \\
\quad Width 384 & 33.02 & 19.73 & 52.75 & $-4.26$ \\
\quad Depth 1 (width 512) & 16.62 & 12.39 & 29.01 & $-28.00$ \\
\quad Depth 12 (width 512) & 26.70 & 16.10 & 42.80 & $-14.21$ \\
\quad Separate, width 512 each & 28.68 & 19.36 & 48.04 & $-8.97$ \\
\quad Separate, parameter-matched & 28.08 & 18.83 & 46.91 & $-10.10$ \\
\bottomrule
\end{tabular}}
\end{table}

\noindent\textbf{Learning objective.} Latent prediction provides the primary
cross-modal signal by preserving and predicting token-level structure.
Pair-distillation is weak in isolation, but improves the joint model by supplying a
global representation-level constraint. Applying both losses to the same content
tokens is counterproductive: latent prediction favors token-specific information,
whereas pair-distillation promotes a shared global summary. We therefore apply
pair-distillation to a dedicated CLS token, isolating its gradients from the tokens used
for latent prediction. Consistent with this interpretation, removing CLS reduces the
joint model to the latent-only level and eliminates nearly all of the pair-distillation
gain.\par
\noindent\textbf{Target construction.} Retrieval is maximized at balanced mixing and
falls sharply when mixing is removed or weakened. Regressing only the pooled teacher
mean yields weak alignment, demonstrating that target equality alone is insufficient
and that the relevant signal lies in the mixed token targets.\par
\noindent\textbf{Pair correspondence.} Random image--text pairing collapses retrieval
despite preserving the target construction. Alignment therefore depends on the shared
semantics of paired samples rather than on the algebra of target mixing alone.\par
\noindent\textbf{Text encoder.} Causal EOS encoding substantially degrades retrieval.
Under causal attention, token $j$ represents only its prefix, while only EOS observes
the complete caption. Token-level cross-modal prediction must therefore infer
order-dependent partial caption states from an image, an underdetermined problem because
each image admits multiple valid captions.\par
\noindent\textbf{Predictor role.} Predictor width is beneficial, while depth has an
intermediate optimum. Direction-specific predictors underperform the shared predictor
even after controlling for parameter count, showing that the gain arises from imposing
one prediction function across both directions rather than from increased capacity.

\section{Conclusion}

U-JEPA demonstrates that cross-modal alignment can emerge from predictive learning without explicitly optimizing image–text similarity through contrastive objectives. After discarding training-only components, the learned encoder representations remain directly comparable across modalities while preserving the inference efficiency of independent dual encoders. Strong performance across retrieval, recognition, and compositional benchmarks demonstrates the transferability of these representations beyond global alignment, with token-level features further enabling fine-grained matching through late interaction. Together, these findings position JEPAs as a promising foundation for general-purpose vision–language representation learning and motivate further investigation at larger scales and across additional paired modalities.

\section{Limitations and Future Work}

The theoretical basis of cross-modal alignment in U-JEPA remains open. Understanding the interplay between prediction, collapse prevention, and representation geometry may clarify when alignment arises. Experiments on CC3M and CC12M at two encoder scales show gains with data and model capacity, but broader studies are needed to characterize scaling and sensitivity to data quality and training duration. Training incurs approximately \(1.7\times\) the per-step computation of contrastive baselines during training. Alternatives such as SIGReg may eliminate teacher forward passes, though their ability to preserve alignment remains untested. Improving training efficiency remains future work. Finally, Winoground reveals weaker image than caption selection, motivating objectives that better preserve fine-grained scene distinctions and relational structure. Due to computational constraints, we leave comprehensive exploration of the hyperparameter to future work.

\clearpage
\section{LLM Usage Statement}
We used large language models, including OpenAI (ChatGPT/GPT), Anthropic (Claude), and Google (Gemini) to assist in refining author-produced content. This assistance covered writing and editing, related work and citations, code and implementation and the presentation of research ideas, hypotheses, claims, and conclusions. We did not use LLMs or generative AI tools for preparing figures or visual materials, or for dataset creation and annotation. We take full responsibility for the accuracy, integrity, and scientific content of this work.
{
    \small
    \bibliographystyle{ieeenat_fullname}
    \bibliography{main}
}

\clearpage

\clearpage
\setcounter{page}{1}
\maketitlesupplementary

\begin{table*}[t]
\centering
\caption{\textbf{CC3M classification and ImageNet robustness.} Accuracy (\%). Zero-shot protocols and dataset metrics match the main classification table; the final column reports ImageNet-1K linear probing (LP).}
\label{tab:supp_cc3m_classification}
\small
\setlength{\tabcolsep}{2.5pt}
\renewcommand{\arraystretch}{1.12}
\resizebox{\textwidth}{!}{%
\begin{tabular}{@{}lrrrrrrrrrrrrrrrrrr@{}}
\toprule
& \multicolumn{11}{c}{\textbf{Zero-shot classification}} & \multicolumn{6}{c}{\textbf{ImageNet zero-shot}} & \textbf{LP} \\
\cmidrule(lr){2-12}\cmidrule(lr){13-18}\cmidrule(lr){19-19}
\textbf{Method}
 & \shortstack{Food\\\cite{bossard2014food101}} & \shortstack{C10\\\cite{krizhevsky2009tiny}} & \shortstack{C100\\\cite{krizhevsky2009tiny}} & \shortstack{SUN\\\cite{xiao2010sun}} & \shortstack{Cars\\\cite{krause2013cars}} & \shortstack{DTD\\\cite{cimpoi2014textures}} & \shortstack{Pets\\\cite{parkhi2012pets}} & \shortstack{Flwr\\\cite{nilsback2008flowers}} & \shortstack{Cal\\\cite{feifei2004learning}} & \shortstack{Euro\\\cite{helber2019eurosat}} & \shortstack{RES\\\cite{cheng2017remote}} & \shortstack{1K\\\cite{deng2009imagenet}} & \shortstack{V2\\\cite{recht2019imagenet}} & \shortstack{A\\\cite{hendrycks2021natural}} & \shortstack{R\\\cite{hendrycks2021faces}} & \shortstack{O\\\cite{hendrycks2021natural}} & \shortstack{Sk.\\\cite{wang2019sketch}} & 1K \\
\midrule
\multicolumn{19}{l}{\textit{ViT-S/16}} \\
CLIP~\cite{radford2021clip} & 11.6 & 50.0 & 22.6 & 32.4 & 0.4 & 11.8 & 4.2 & 9.6 & 42.7 & 12.8 & \textbf{20.4} & 16.6 & 13.8 & 3.7 & 18.1 & 22.9 & 6.2 & 50.8 \\
U-JEPA & \textbf{12.6} & \textbf{55.5} & \textbf{24.1} & \textbf{33.1} & \textbf{1.0} & \textbf{16.9} & \textbf{10.2} & \textbf{12.9} & \textbf{51.7} & \textbf{20.3} & 17.4 & \textbf{19.2} & \textbf{16.4} & \textbf{4.2} & \textbf{23.9} & \textbf{27.3} & \textbf{10.1} & \textbf{54.2} \\
\midrule
\multicolumn{19}{l}{\textit{ViT-B/16}} \\
CLIP~\cite{radford2021clip} & 11.1 & 47.6 & 22.0 & 34.0 & \textbf{0.9} & 12.4 & 11.8 & 11.3 & 44.3 & 11.9 & 18.2 & 17.1 & 14.4 & 3.2 & 16.9 & 20.9 & 5.9 & 52.9 \\
U-JEPA & \textbf{12.9} & \textbf{57.7} & \textbf{31.7} & \textbf{35.1} & 0.5 & \textbf{19.3} & \textbf{13.2} & \textbf{13.7} & \textbf{56.5} & \textbf{18.9} & \textbf{22.4} & \textbf{22.0} & \textbf{18.6} & \textbf{5.1} & \textbf{30.0} & \textbf{29.1} & \textbf{13.2} & \textbf{58.9} \\
\bottomrule
\end{tabular}}
\end{table*}

\section{Experimental Setup}
\label{sec:supp_experimental_setup}

\begin{table}[t]
\centering
\caption{\textbf{Text backbone.} Shared across U-JEPA configurations.}
\label{tab:supp_text_backbone}
\small
\setlength{\tabcolsep}{4pt}
\begin{tabular*}{\columnwidth}{@{\extracolsep{\fill}}ll@{}}
\toprule
\textbf{Setting} & \textbf{Specification} \\
\midrule
Architecture & Bidirectional Encoder \\
Layers & 12 \\
Hidden width & 512 \\
Attention heads & 8 \\
MLP expansion ratio & 4 \\
Tokenizer & CLIP BPE \\
Context length & 77 \\
Vocabulary (including padding) & 49,409 \\
\bottomrule
\end{tabular*}
\end{table}

\begin{table}[t]
\centering
\caption{\textbf{Vision backbones.} U-JEPA Small and Base configurations.}
\label{tab:supp_vision_backbones}
\small
\setlength{\tabcolsep}{4pt}
\begin{tabular*}{\columnwidth}{@{\extracolsep{\fill}}lcc@{}}
\toprule
\textbf{Setting} & \textbf{ViT-S/16} & \textbf{ViT-B/16} \\
\midrule
Layers & 12 & 12 \\
Hidden width & 384 & 768 \\
Attention heads & 6 & 12 \\
MLP expansion ratio & 4 & 4 \\
Image resolution & $224\times224$ & $224\times224$ \\
Patch size & $16\times16$ & $16\times16$ \\
Patch tokens & 196 & 196 \\
\bottomrule
\end{tabular*}
\end{table}

Backbone specifications are given in Tables~\ref{tab:supp_text_backbone}
and~\ref{tab:supp_vision_backbones}. The visual output is projected to 512
dimensions.

\begin{table*}[t]
\centering
\caption{\textbf{U-JEPA pretraining settings.} Shared entries apply to all three configurations.}
\label{tab:supp_pretraining_hyperparameters}
\small
\setlength{\tabcolsep}{6pt}
\renewcommand{\arraystretch}{1.08}
\begin{tabular*}{\textwidth}{@{\extracolsep{\fill}}lccc@{}}
\toprule
\textbf{Setting} & \textbf{Small CC3M} & \textbf{Base CC3M} & \textbf{Base CC12M} \\
\midrule
Predictor layers & 2 & 4 & 4 \\
Predictor width / heads & 512 / 4 & 512 / 4 & 512 / 4 \\
Global batch size & 1,024 & 1,024 & 4,096 \\
GPUs & 4 & 4 & 8 \\
Batch per GPU & 256 & 256 & 512 \\
Gradient accumulation steps & 1 & 1 & 1 \\
\midrule
Training epochs & \multicolumn{3}{c}{60} \\
Optimizer & \multicolumn{3}{c}{AdamW} \\
Optimizer $(\beta_1,\beta_2)$ / $\epsilon$ & \multicolumn{3}{c}{$(0.9,0.999)$ / $10^{-8}$} \\
Precision & \multicolumn{3}{c}{BF16} \\
Learning rate (initial / peak / final) & \multicolumn{3}{c}{$2\times10^{-4}$ / $5\times10^{-4}$ / $5\times10^{-4}$} \\
Learning-rate schedule & \multicolumn{3}{c}{5-epoch linear warmup, then constant} \\
Weight decay & \multicolumn{3}{c}{Cosine increase from 0.04 to 0.4} \\
Teacher EMA momentum & \multicolumn{3}{c}{Linear increase from 0.996 to 1.0} \\
Minimum crop scale / horizontal flip probability & \multicolumn{3}{c}{0.5 / 0} \\
Joint embedding dimension & \multicolumn{3}{c}{512} \\
Image context & \multicolumn{3}{c}{49 random tokens, selected after encoding} \\
Text context & \multicolumn{3}{c}{All 77 positions, including padding} \\
Image targets & \multicolumn{3}{c}{4 random blocks of 30 patches} \\
Text targets & \multicolumn{3}{c}{4 random spans of 4 tokens} \\
Token regression / loss weight & \multicolumn{3}{c}{Smooth-$L_1$ / 1} \\
Mean / summary injection & \multicolumn{3}{c}{1 / 1} \\
Prototype count  & \multicolumn{3}{c}{1,024} \\
Pair-distillation loss weight & \multicolumn{3}{c}{0.1} \\
Student / teacher temperature & \multicolumn{3}{c}{0.1 / 0.04} \\
Prototype logit center momentum & \multicolumn{3}{c}{0.9} \\
\bottomrule
\end{tabular*}
\end{table*}

All U-JEPA models are trained from scratch using the settings in
Table~\ref{tab:supp_pretraining_hyperparameters}. One predictor serves both
prediction directions. Biases and one-dimensional parameters are excluded
from weight decay.

\subsection{CLIP Baselines}
\label{sec:clip_baselines}

\begin{table}[t]
\centering
\caption{\textbf{Long-schedule CLIP training settings.}}
\label{tab:supp_clip_hyperparameters}
\small
\setlength{\tabcolsep}{4pt}
\begin{tabular*}{\columnwidth}{@{\extracolsep{\fill}}ll@{}}
\toprule
\textbf{Setting} & \textbf{Specification} \\
\midrule
Image backbone & ViT-B/16 \\
Text backbone & Native CLIP, causal, EOS readout \\
Image resolution / text context & 224 / 77 \\
Epochs & 105 \\
Global batch / GPUs & 4,096 / 8 \\
Batch per GPU / accumulation & 512 / 1 \\
Optimizer & AdamW \\
Learning rate & $5\times10^{-4}$ \\
Warmup / subsequent schedule & 10,000 steps / cosine decay \\
Weight decay & 0.2 \\
Optimizer $(\beta_1,\beta_2)$ / $\epsilon$ & $(0.9,0.98)$ / $10^{-6}$ \\
Precision & BF16 \\
\bottomrule
\end{tabular*}
\end{table}

Our CLIP baselines~\cite{radford2021clip} are trained from scratch on
CC3M and CC12M at the corresponding image-backbone scales. The 105-epoch
CLIP configuration is listed in Table~\ref{tab:supp_clip_hyperparameters}.
We also evaluate released CC12M OpenCLIP~\cite{ilharco2021openclip,cherti2023reproducible}
and PowerCLIP-S~\cite{kawamura2026powerclip} checkpoints.

\section{Evaluation Setup}
\label{sec:supp_evaluation}

\subsection{Shared Readouts and Preprocessing}
\label{sec:supp_eval_readouts}

U-JEPA uses the frozen EMA target encoders, without the predictor or prototype
head. Global similarity is cosine similarity between $\ell_2$-normalized
image-patch and valid-text-token means. Text features include BOS/EOS and
exclude padding. Locally evaluated baselines retain their native global
readouts; published references retain their original protocols.

For unit-normalized patch features $u_p$ and text features $v_t$, MaxSim is
\begin{align}
 S_T(I,C) &= \frac{1}{|C|}\sum_{t\in C}\max_{p\in I}v_t^\top u_p,
 \label{eq:supp_maxsim_text}\\
 S_I(I,C) &= \frac{1}{|I|}\sum_{p\in I}\max_{t\in C}u_p^\top v_t.
 \label{eq:supp_maxsim_image}
\end{align}
Padding and image summary tokens are excluded. Query-side retrieval uses
$S_T$ for text-to-image and $S_I$ for image-to-text. Caption discrimination
uses $S_T$, including the fixed-readout Winoground and sDCI results.

\subsection{Zero-Shot Image--Text Retrieval}
\label{sec:supp_eval_retrieval}

MS-COCO~\cite{lin2014coco} uses the Karpathy 5K test split: 5,000 images and
all 25,010 captions in the local annotations. Flickr30K~\cite{young2014flickr}
uses the 1K test split: 1,000 images and 5,000 captions. Each query ranks the
complete opposite-modality gallery. A caption has one correct image; an
image may match any of its associated captions. We report R@1/5/10
separately in each direction and their bidirectional R@1 sum. Mean and
query-side MaxSim use identical checkpoints and galleries, without reranking.

\subsection{Zero-Shot Classification}
\label{sec:supp_eval_classification}

We evaluate Food-101, CIFAR-10/100, SUN397, Stanford Cars, DTD, Oxford-IIIT
Pets, Flowers-102, Caltech-101, EuroSAT, RESISC45, and ImageNet-1K. We use test
splits, except ImageNet-1K, which uses its 50,000-image validation set.
CLIP-benchmark datasets use the supplied test shards, class ordering, names,
and templates; other datasets use the stored benchmark names and ensembles.

Each dataset-specific template is instantiated with a class name and encoded.
Prompt embeddings are individually $\ell_2$-normalized, averaged within each
class, and normalized again. Classification selects the class prototype with
maximum cosine similarity to the image embedding. Pets, Flowers-102, and
Caltech-101 report mean-per-class accuracy; the remaining datasets report
sample-weighted top-1 accuracy.

\subsection{ImageNet Distribution Shifts}
\label{sec:supp_eval_imagenet_shifts}

We reuse the ImageNet-1K prototypes and prompts on the complete
ImageNet-V2 matched-frequency (10,000 images), ImageNet-A (7,500),
ImageNet-R (30,000), ImageNet-O (2,000), and ImageNet-Sketch (50,889) splits.
V2 numeric labels map directly to ImageNet-1K indices. A/R/O predictions are
restricted to their respective 200-class subsets using CLIP-benchmark
WordNet mappings; V2 and Sketch use all 1,000 classes. All columns report
top-1 accuracy. ImageNet-O is evaluated as restricted-label classification,
rather than out-of-distribution detection.

\subsection{ImageNet Linear Probing}
\label{sec:supp_eval_linear_probe}

The probe uses all 1,281,167 ImageNet-1K training images and 50,000 validation
images. U-JEPA applies non-affine LayerNorm to each output patch token before
mean pooling; contrastive controls apply non-affine LayerNorm to their native
projected image vector. Features are standardized using training-only means
and standard deviations, with standard deviations clamped at $10^{-6}$.
The encoder is frozen, and only a linear classifier with bias is trained.
Table~\ref{tab:supp_probe_settings} gives the optimization settings.
Reported accuracy is the mean over seeds 0, 1, and 2.

\begin{table}[t]
\centering
\caption{\textbf{Frozen-probe optimization.} Dense settings apply to ADE20K
and COCO-Stuff.}
\label{tab:supp_probe_settings}
\small
\setlength{\tabcolsep}{3pt}
\renewcommand{\arraystretch}{1.12}
\begin{tabular}{@{}p{0.28\columnwidth}p{0.24\columnwidth}p{\dimexpr0.48\columnwidth-4\tabcolsep\relax}@{}}
\toprule
\textbf{Setting} & \textbf{ImageNet} & \textbf{Dense} \\
\midrule
Head & Linear & Patchwise linear \\
Optimizer & AdamW & AdamW \\
Epochs & 40 & 40 \\
Batch size & 1,024 & 256 \\
Learning rate & $10^{-3}$ & \shortstack[l]{$10^{-4},\ 10^{-3},$\\$10^{-2}$} \\
Weight decay & $10^{-4}$ & $10^{-4}$ \\
Schedule & Cosine & Linear warm up + cosine \\
Warmup epochs & 0 & 1 \\
\bottomrule
\end{tabular}
\end{table}

\subsection{SugarCrepe}
\label{sec:supp_eval_sugarcrepe}

We evaluate all 7,511 SugarCrepe~\cite{hsieh2023sugarcrepe} examples across
add-object/attribute, replace-object/attribute/relation, and
swap-object/attribute. Each positive caption must score strictly above its
foil; ties fail. Object and attribute scores are unweighted means over their
respective add, replace, and swap subsets. Relation uses replace-relation.
Overall accuracy weights all examples equally.

\subsection{ARO}
\label{sec:supp_eval_aro}

ARO~\cite{yuksekgonul2023when} comprises VG-Attribution (28,748 examples),
VG-Relation (23,937), COCO-Order (25,010), and Flickr30K-Order (5,000).
Visual Genome bounding-box crops are applied before shared preprocessing.
The positive caption must strictly outscore its negative. VG-Relation is
macro-averaged over relation classes; VG-Attribution is macro-averaged over
attribute-pair groups with at least 25 examples.

Order tasks compare the correct caption with four perturbations generated
using the benchmark functions and seed 1. All models use the same candidate
sets. Accuracy is five-way argmax accuracy; the positive is first in the
candidate list, and the first maximum wins ties.

\subsection{VALSE}
\label{sec:supp_eval_valse}

We use the 7,702 human-validated VALSE~\cite{parcalabescu2022valse} examples
with at least two annotator votes supporting the original caption. Correctness
requires positive-caption similarity strictly above foil similarity. We
report per-piece accuracies and their macro-average. Example-weighted overall
accuracy and the macro-average excluding the Foil-it object-recognition
control are recorded separately.

\subsection{Winoground}
\label{sec:supp_eval_winoground}

All 400 Winoground~\cite{thrush2022winoground} groups are evaluated, each
with two images and two captions. For the $2\times2$ score matrix $s_{ij}$,
text correctness requires $s_{00}>s_{01}$ and $s_{11}>s_{10}$; image
correctness requires $s_{00}>s_{10}$ and $s_{11}>s_{01}$. Group correctness
requires both. Metrics are percentages of groups satisfying these strict
conditions. The main-table MaxSim results use one $S_T$ matrix for all
three metrics.

\subsection{Frozen Dense Semantic Segmentation}
\label{sec:supp_eval_segmentation}

ADE20K~\cite{zhou2017ade20k} uses 20,210 training images, 2,000 validation
images, and 150 classes. COCO-Stuff 2017~\cite{caesar2018cocostuff} uses
118,287 training images, 5,000 validation images, and 171 classes after
mapping the official category IDs. Void labels are ignored.

We extract frozen ViT-B/16 final backbone-normalized patch features before
joint-space projection: 768 channels on the native $14\times14$ grid,
excluding the summary token. Entire RGB images are bicubically resized to
$224\times224$ and use identical ImageNet normalization for all models.
Features are standardized per channel using training patches only, with
standard deviations clamped at $10^{-6}$. There is no training augmentation.

A single patchwise linear layer maps 768 channels to 150 or 171 classes.
Training labels are nearest-neighbor resized to $14\times14$; optimization
follows Table~\ref{tab:supp_probe_settings}. Validation logits are bilinearly
upsampled to the original mask resolution with
\texttt{align\_corners=False} before argmax. We report validation mIoU and
mean per-class pixel accuracy (mAcc), averaged across three seeds.

\subsection{Single-Summary Dense-Caption Matching}
\label{sec:supp_eval_sdci}

sDCI~\cite{urbanek2024picture} uses the authors' ordered batch-eight
protocol. All contains 1,382 complete-image and region-crop pairs; Base
contains 108 complete-image pairs. Each pair uses its first supplied
summary, truncated to the model's context length. Crops precede CLIP's
shortest-edge bicubic resize to 224 and $224\times224$ center crop.
Checkpoint-native normalization is retained.

Pairs preserve official test-image order and within-image annotation
insertion order. Contiguous batches contain eight pairs, without shuffling;
the final partial batch is retained. Matching (M) is image-to-caption argmax
accuracy among the batch's positive summaries; the first maximum wins ties.
Negative discrimination (N) requires the positive to strictly outscore its
designated foil.

All and Base use the first supplied swap negative. Hardest retains all
1,382 pairs and selects the highest supplied reference-CLIP-scoring negative
across basic, layout, and swap candidates. Thus, Hardest changes N but not M.
U-JEPA uses mean pooling or $S_T$. LeVLJEPA retains the authors' two
predictor paths and native score combination for this benchmark.

\subsection{Contrastive Fine-Tuning on COCO}
\label{sec:supp_eval_coco_finetune}

Both towers are fine-tuned on COCO train with symmetric InfoNCE, sampling one
caption per image. Shared settings are batch size 512, AdamW learning rate
$2\times10^{-5}$, weight decay 0.1, gradient clipping at norm 1, and seed 0.
The 1-, 5-, and 10-epoch schedules are independent runs, each with 5\%
linear warmup and cosine decay, evaluated at its final epoch.

OpenCLIP retains its pretrained logit scale; predictive methods initialize
temperature at 0.07. Readouts are U-JEPA mean pooling, OpenCLIP native
projected embeddings, and LeVLJEPA pre-predictor projections. Retrieval uses
the complete COCO 5K test gallery. Query-side MaxSim is evaluated separately
on the same fine-tuned weights. Adaptation results use one seed.

\subsection{Additional Readout Audit}
\label{sec:supp_eval_readout_audit}

The separate Winoground query-side audit uses $S_I$ for text correctness
and $S_T$ for image correctness; group correctness requires both. These
results are distinct from the fixed-$S_T$ main-table protocol.

\begin{figure*}[!t]
\centering
\includegraphics[width=0.78\textwidth]{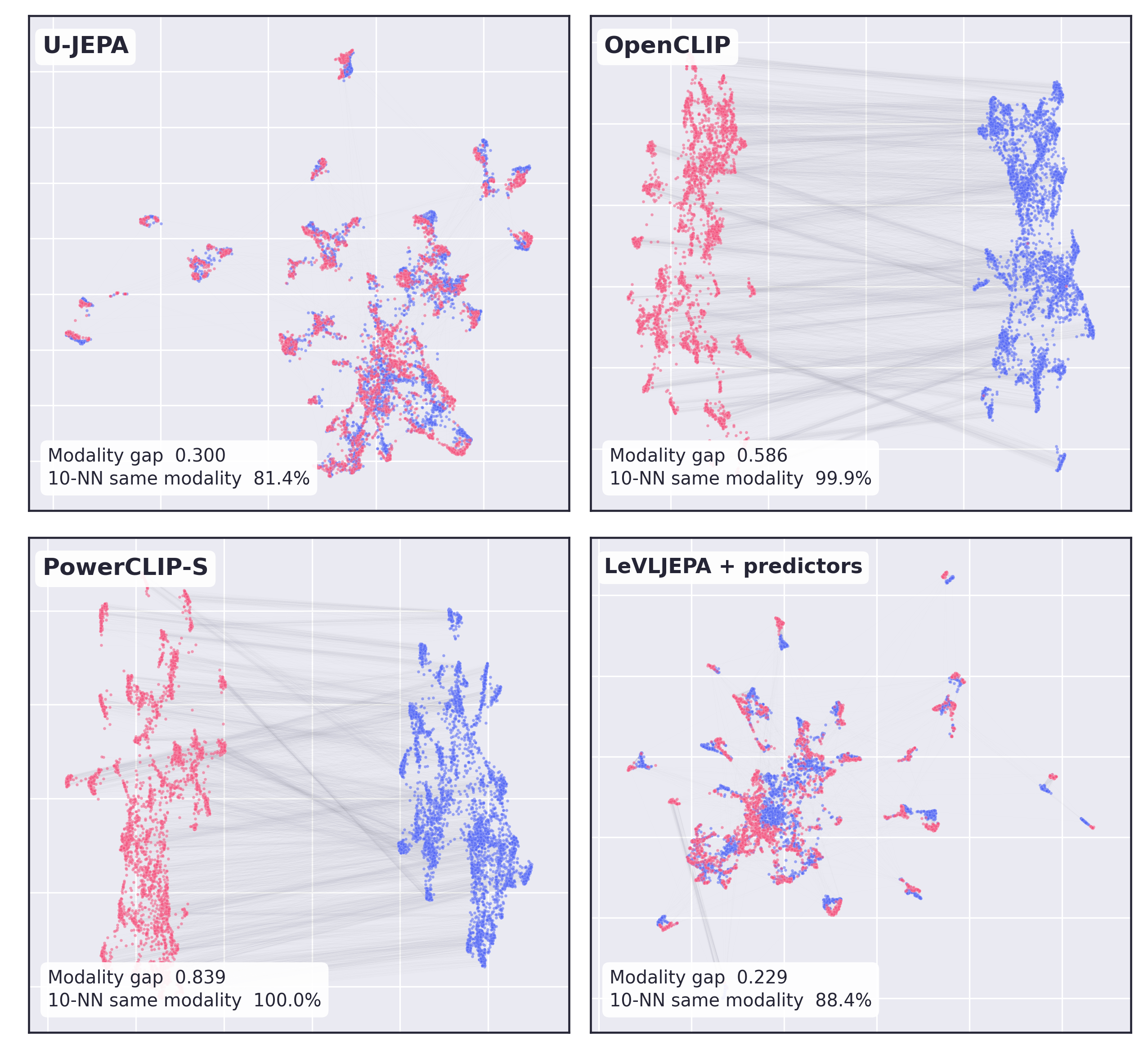}
\caption{\textbf{Modality geometry on COCO.} Blue denotes images and pink
text; thin lines connect matched pairs. Independently fitted UMAP projections
use identical settings and all 10,000 embeddings per model; the same seeded
subset of 1000 pairs is displayed in each panel. Numeric annotations use all
embeddings in the original latent space. PowerCLIP-S uses the released SAM2
checkpoint; LeVLJEPA retains both predictors to match its retrieval readout.}
\label{fig:supp_modality_gap}
\end{figure*}

\section{Additional Results}
\label{sec:supp_additional_results}

\subsection{Contrastive Fine-Tuning on COCO}
\label{sec:supp_results_coco_finetune}

Under the protocol in Sec.~\ref{sec:supp_eval_coco_finetune}, OpenCLIP
achieves its highest pooled R@1 sum after one epoch (88.90), while U-JEPA
peaks after five epochs (87.13; Table~\ref{tab:supp_coco_finetune}). At that
five-epoch checkpoint, query-side MaxSim increases U-JEPA's R@1 sum to
89.23, showing that its late-interaction benefit persists after contrastive
fine-tuning.

\begin{table}[t]
\centering
\caption{\textbf{COCO contrastive fine-tuning.} R@1 (\%) on the Karpathy
5K test split. Each schedule is an independent run. MS denotes query-side
MaxSim at the checkpoint with the highest pooled R@1 sum among the tested
schedules: epoch 1 for OpenCLIP, 10 for LeVLJEPA, and 5 for U-JEPA.}
\label{tab:supp_coco_finetune}
\small
\setlength{\tabcolsep}{3pt}
\renewcommand{\arraystretch}{1.08}
\begin{tabular*}{\columnwidth}{@{\extracolsep{\fill}}llrr@{}}
\toprule
Method & Schedule & T$\to$I & I$\to$T \\
\midrule
OpenCLIP  & Zero-shot & 26.89 & 36.18 \\
 & 1 epoch & 36.50 & 52.40 \\
 & 5 epochs & 36.24 & 51.28 \\
 & 10 epochs & 35.14 & 47.86 \\
 & MS (1 epoch) & 0.26 & 3.86 \\
\midrule
LeVLJEPA & Zero-shot & 0.03 & 0.04 \\
 & 1 epoch & 4.36 & 6.16 \\
 & 5 epochs & 16.16 & 21.96 \\
 & 10 epochs & 19.84 & 25.84 \\
 & MS (10 epochs) & 4.34 & 1.36 \\
\midrule
U-JEPA & Zero-shot & 23.88 & 37.50 \\
 & 1 epoch & 34.23 & 45.48 \\
 & 5 epochs & 37.87 & 49.26 \\
 & 10 epochs & 36.46 & 48.10 \\
 & MS (5 epochs) & 39.35 & 49.88 \\
\bottomrule
\end{tabular*}
\end{table}

\subsection{Modality Gap}
\label{sec:supp_modality_gap}

We compare modality geometry on 5,000 COCO test images and one caption per
image, using each model's retrieval readout. LeVLJEPA's two normalized
prediction paths are concatenated in the corresponding image/text order
so that their dot product reproduces its retrieval score. LeVLJEPA uses
DataComp-L; the remaining models use CC12M.

We quantify the modality gap using the Euclidean distance between the
centroids of unit-normalized image and text embeddings,
$g = \|\bar{z}_I - \bar{z}_T\|_2$~\cite{liang2022mindthegap}.
We further assess local modality segregation by measuring the fraction
of same-modality embeddings among each embedding's ten nearest neighbors,
excluding itself. In our balanced sample, a fraction of $50\%$ indicates
modality mixing, whereas $100\%$ indicates complete local segregation.
Both metrics are computed directly in the original embedding space,
without additional centering or learned transformations.
We interpret the modality gap as a geometric property of the learned
representations, highlighting differences from the embedding geometry
induced by contrastive learning. We do not draw conclusions about its
implications for downstream task performance, which remain an important
direction for future investigation.

\begin{table}[!htb]
\centering
\caption{\textbf{Modality geometry on 5,000 COCO pairs.} Gap: centroid
distance; 10-NN: same-modality neighbor fraction (\%).
$^\ast$LeVLJEPA retains both predictors and uses DataComp-L; other models use CC12M.}
\label{tab:supp_modality_gap_quantitative}
\small
\setlength{\tabcolsep}{3pt}
\renewcommand{\arraystretch}{1.12}
\begin{tabular*}{\columnwidth}{@{\extracolsep{\fill}}lrr@{}}
\toprule
Method & Gap & 10-NN \\
\midrule
OpenCLIP & 0.586 & 99.9 \\
PowerCLIP-S & 0.839 & 100.0 \\
LeVLJEPA$^\ast$ & 0.229 & 88.4 \\
U-JEPA & 0.300 & 81.4 \\
\bottomrule
\end{tabular*}
\end{table}

Figure~\ref{fig:supp_modality_gap} visualizes the modality geometry.
U-JEPA has fewer same-modality neighbors than the contrastive references
(81.4\% versus at least 99.9\%; Table~\ref{tab:supp_modality_gap_quantitative}).
LeVLJEPA has a smaller centroid gap but lower retrieval accuracy, so centroid
gap alone does not determine retrieval performance.

\begin{figure*}[p]
\centering
\includegraphics[width=\textwidth,height=0.80\textheight,keepaspectratio]{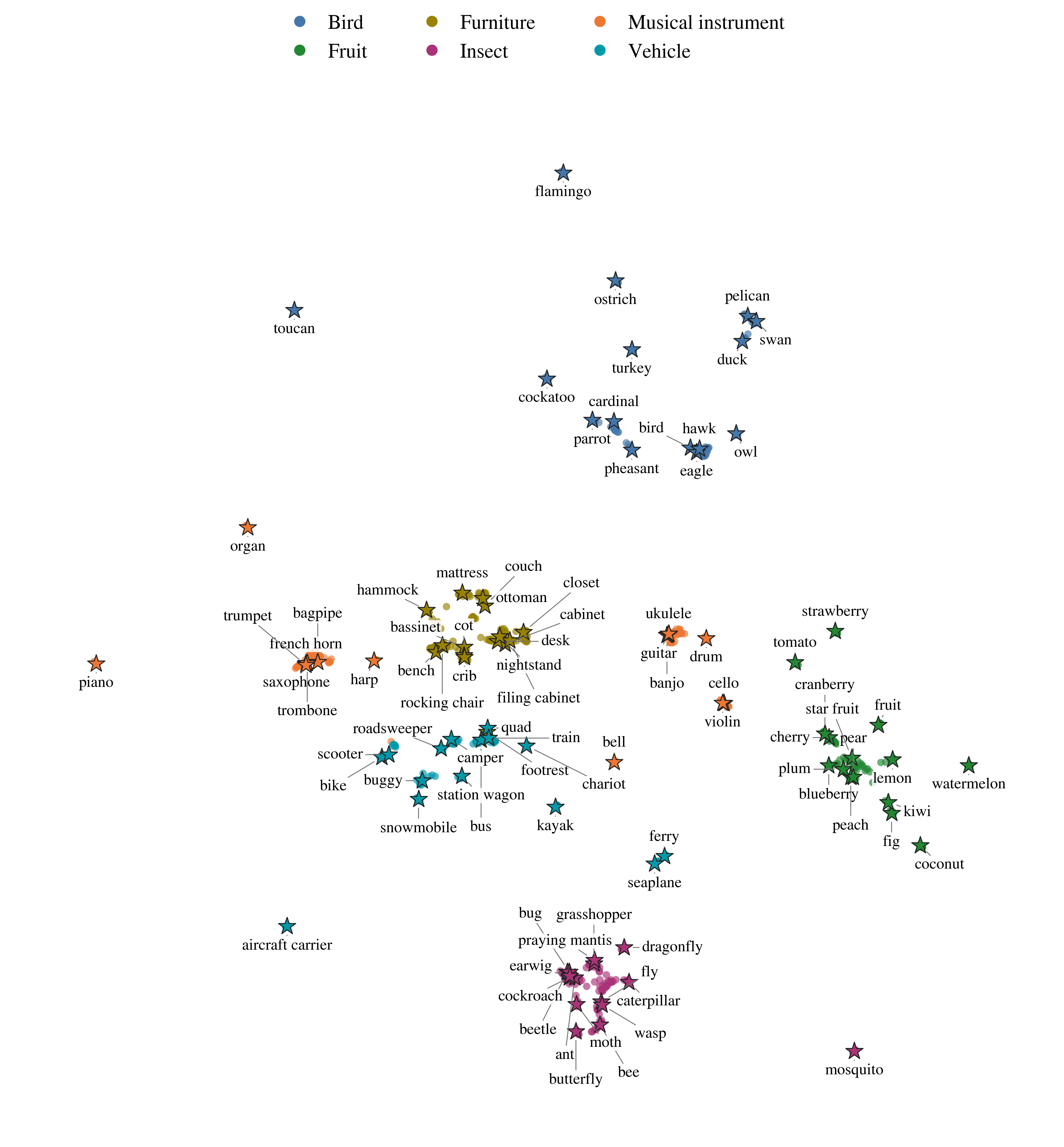}
\caption{\textbf{Concept organization in U-JEPA.}
Joint UMAP of 900 THINGS images and 90 text prototypes from six
THINGSplus categories~\cite{stoinski2024thingsplus} (15 concepts per category,
ten images per concept; seed 23). Text prototypes average four prompts (\texttt{a photo of}, \texttt{a picture of},
\texttt{an image of}, \texttt{a close-up of}, followed by the concept name);
embeddings and averaged prototypes are $\ell_2$-normalized.}
\label{fig:supp_concept_organization}
\end{figure*}

\begin{figure*}[p]
\centering
\includegraphics[width=\textwidth,height=0.80\textheight,keepaspectratio]{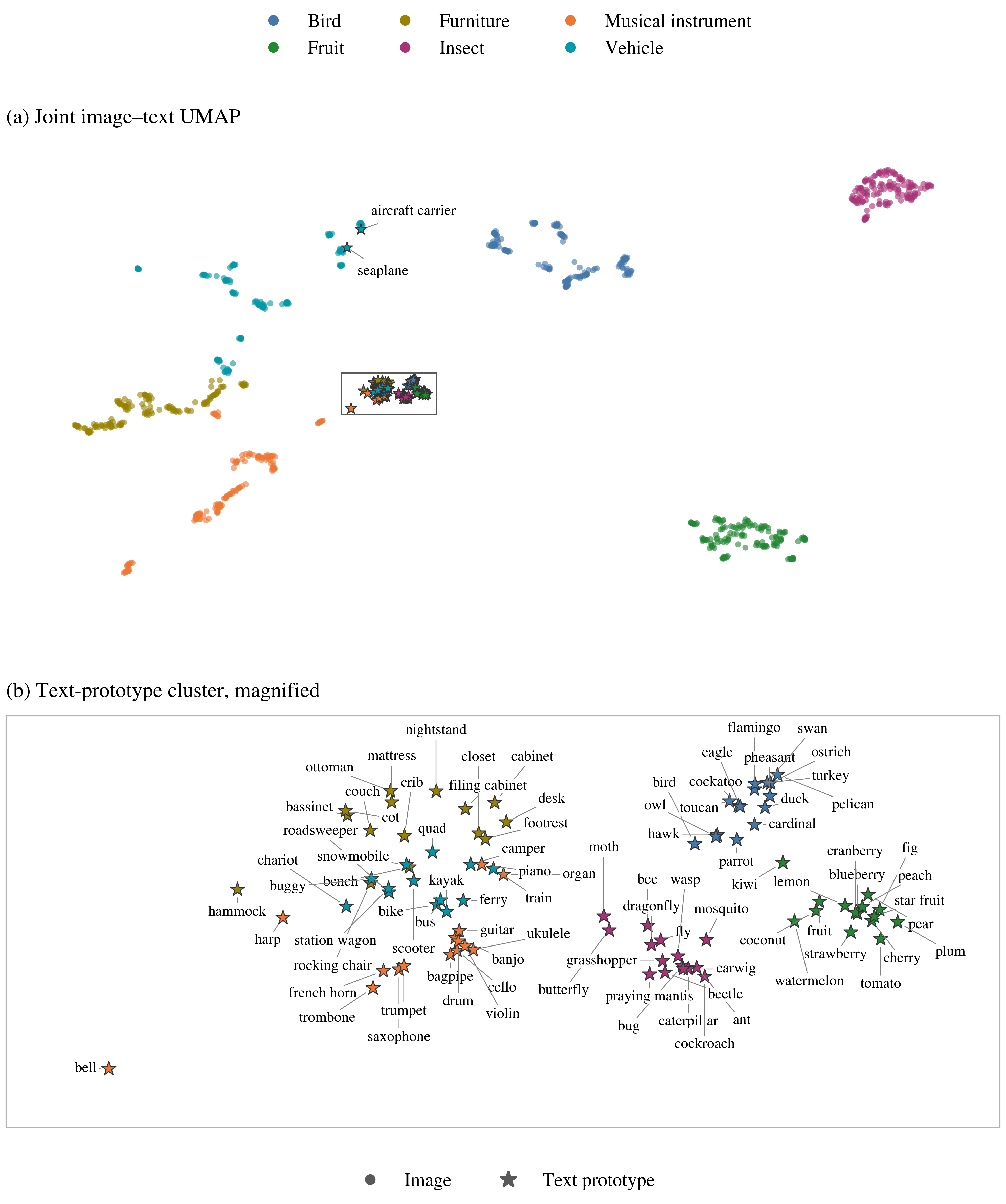}
\caption{\textbf{Concept organization in OpenCLIP.}
OpenCLIP checkpoint evaluated on the
same 900 images and 90 concepts as Figure~\ref{fig:supp_concept_organization}.
We use projected image CLS and text EOT readouts with native CLIP
normalization; prompts, prototype averaging, and $\ell_2$ normalization
follow the U-JEPA figure. \textbf{(a)} Joint UMAP of image embeddings
(circles) and text prototypes (stars), using the same settings as U-JEPA
but fitted independently. \textbf{(b)} Magnification of the boxed text
cluster, retaining the original UMAP coordinates. Its 88 prototypes are
labeled here; the two outlying prototypes are labeled in (a).
Colors indicate the six THINGSplus categories. Labels are used only for
display. Global distances and axes across the independently fitted
projections are not directly comparable.}
\label{fig:supp_concept_organization_openclip}
\end{figure*}

Figures~\ref{fig:supp_concept_organization}
and~\ref{fig:supp_concept_organization_openclip} compare concept organization
on the same 900-image, 90-concept THINGSplus subset. In U-JEPA, images and
their text prototypes generally occupy shared concept neighborhoods, with
several related concepts grouped nearby. OpenCLIP forms semantic image
clusters, while most text prototypes occupy a separate compact region;
the magnified view reveals semantic grouping within that text region.
Both models retain concept discrimination: nearest-text-prototype accuracy
in the original embedding space is 80.9\% for U-JEPA and 80.7\% for OpenCLIP.
Together with the COCO neighborhood statistics, this comparison supports
reduced modality segregation in U-JEPA alongside retained semantic
structure. The independently fitted UMAP projections provide qualitative
evidence; their axes and global distances are not directly comparable.

\clearpage
\clearpage

\clearpage

\end{document}